\documentclass[11pt,fleqn,twoside,titlepage]{cslarticle}
\newcommand{\ifelse}[2]{#2}
 
\cslreportnumber{\ifelse{}{CSL} Technical Report \ifelse{for CLARISSA Project}{SRI-CSL-2024-01}}
\ifelse{\acknowledge{SRI Project 100651 under subcontract to Honeywell
in support of AFRL and DARPA ARCOS Program,
Contract FA8750-20-C-0512.
}}
{\acknowledge{\vspace*{-2ex}SRI Project 100651 under subcontract to Honeywell
in support of AFRL and DARPA ARCOS Program,
Contract FA8750-20-C-0512.\\
Distribution Statement “A” (Approved for Public Release, Distribution Unlimited).
}}

\newcounter{ofootnote}

\newcommand{\atwo}{Assurance 2.0}
\newcommand{\cl}{{\sc Clarissa}}
\newcommand{\cla}{\ifelse{{\sc Clarissa}}{Assurance 2.0}}

\newcommand{\clasce}{{\sc Clarissa/asce}}
\ifelse{\newcommand{\CL}{C{\smaller LARISSA}}}{}
\ifelse{}{}

\ifelse{}{} 
\newcommand{\footnotet}{\ifelse{\footnote}{\comment}}

\newcommand{\vbar}{\,|\,}

\usepackage[bookmarks=true,hyperfigures=true,colorlinks=true,linkcolor=blue,citecolor=blue,backref=page,pagebackref=false,pdfpagemode=fullscreen,plainpages=false,pdfpagelabels]{hyperref}
\usepackage[all]{hypcap}
\usepackage{wrapfig}
\usepackage[normalem]{ulem} 
\usepackage{caption,cite,url,relative,latexsym,alltt}
\usepackage[utf8]{inputenc}
\DeclareUnicodeCharacter{25A1}{\ensuremath\Box}
\DeclareUnicodeCharacter{25C7}{\ensuremath\Diamond}
\DeclareSymbolFont{symbolsC}{U}{txsyc}{m}{n}
\DeclareMathSymbol{\strictif}{\mathrel}{symbolsC}{74}

\def\BigLaTeX{{\rm L\kern-.36em\raise.3ex\hbox{\smaller\smaller A}\kern-.15em
    T\kern-.1667em\lower.7ex\hbox{E}\kern-.125emX}}
\def\BoldLaTeX{{\bf L\kern-.36em\raise.3ex\hbox{\smaller\smaller\bf A}\kern-.15em
    T\kern-.1667em\lower.7ex\hbox{E}\kern-.125emX}}
\def\BibTeX{{\rm B\kern-.05em{\sc i\kern-.025em b}\kern-.08em
    T\kern-.1667em\lower.7ex\hbox{E}\kern-.125emX}}

\newlength{\hsbw}
\def\extrawidth{0.5in}

\newcounter{sessioncount}
\newenvironment{session*}{\begin{flushleft}
 \refstepcounter{sessioncount}
 \setlength{\hsbw}{\linewidth}
 \addtolength{\hsbw}{-\arrayrulewidth}
 \addtolength{\hsbw}{-\tabcolsep}
 \begin{tabular}{@{}|c@{}|@{}}\hline 
 \begin{minipage}[b]{\hsbw}
 \vspace*{-.5pt}
 \begin{flushright}
 \rule{0.01in}{.15in}\rule{0.3in}{0.01in}\hspace{-0.35in}
 \raisebox{0.04in}{\makebox[0.3in][c]{\footnotesize \thesessioncount}}
 \end{flushright}
 \vspace*{-.57in}
 \begingroup\small\vspace*{1.0ex}\begin{alltt}}{\end{alltt}\endgroup\end{minipage}\\ \hline 
 \end{tabular}
 \end{flushleft}}
\def\sessionsize{\small}
\def\smallsessionsize{\small}

\newcommand{\exmemo}[1]{}
\newcommand{\memo}[1]{\mbox{}\par\vspace{0.25in}
\setlength{\hsbw}{\linewidth}
\addtolength{\hsbw}{-2\fboxsep}
\addtolength{\hsbw}{-2\fboxrule}
\noindent\fbox{\parbox{\hsbw}{{\bf Memo: }#1}}\vspace{0.25in}}

\newcommand{\mem}[2]{\mbox{}\par\vspace{0.25in}
\setlength{\hsbw}{\linewidth}
\addtolength{\hsbw}{-2\fboxsep}
\addtolength{\hsbw}{-2\fboxrule}
\noindent\fbox{\parbox{\hsbw}{{\bf #1: }#2}}\vspace{0.25in}}
\newcommand{\comment}[1]{}

\newcommand{\exfootnote}[1]{}
\newlength{\sblen}
\newlength{\overhang}

\def\SetFigFont#1#2#3{\rm}

\newcommand{\excite}[1]{}

\newcommand{\arxiv}[1]{\href{https://arxiv.org/abs/#1}{\tt arXiv:#1}}

\renewcommand{\memo}[1]{}
\renewcommand{\mem}[2]{}

\title{Defeaters and Eliminative Argumentation \ifelse{in \CL}{\newline In Assurance 2.0}\\[1ex]
\large
\ifelse{Distribution Statement “A” (Approved for Public Release, Distribution Unlimited)
}{}
}
\author{Robin Bloomfield (Adelard, part of NCC Group, and City, Univ.\
of London),\\Kate Netkachova (Adelard), and John Rushby (SRI)
\\[1ex]As Members of the  C{\small LARISSA} Team\\
\emph{\smaller Honeywell, Adelard, UT Dallas, and SRI}
\ifelse{}{\\[1ex]Also issued as a  C{\small LARISSA} Technical Report under the title\\
Defeaters and Eliminative Argumentation in C{\small LARISSA}}
}
\begin{document}
\maketitle

\begin{abstract}

\begin{quotation}
A traditional assurance case employs a positive argument in which
reasoning steps, grounded on evidence and assumptions, sustain a top
claim that has external significance.  Human judgement is required to
check the evidence, the assumptions, and the narrative justifications for
the reasoning steps; if all are assessed good, then the top claim can
be accepted.

A valid concern about this process is that human judgement is fallible
and prone to confirmation bias.  The best defense against this concern
is vigorous and skeptical debate and discussion in the manner of a
dialectic or Socratic dialog.  There is merit in recording aspects of
this discussion for the benefit of subsequent developers and
assessors.  \emph{Defeaters} are a means doing this: they express
doubts about aspects of the argument and can be developed into
subcases that confirm or refute the doubts, and can record them as
documentation to assist future consideration.

This report describes how defeaters, and multiple levels of defeaters,
should be represented and assessed in \atwo\ and its \clasce\ tool
support.  These mechanisms also support \emph{eliminative
argumentation}, which is a contrary approach to assurance, favored by
some, that uses a negative argument to refute all reasons why the top
claim could be false.

\end{quotation}

\end{abstract}

\newpage

\tableofcontents

\listoffigures

\cleardoublepage

\section{Introduction to Defeaters}

This report assumes general familiarity with Assurance Cases
\cite{Rushby:Cases15}, and with Assurance 2.0
\cite{Bloomfield&Rushby:Assurance2} and its \cl\ toolset
\cite{Varadarajan-all:DASC24} in particular.

The primary criterion for a satisfactory assurance case in the \atwo\
methodology \ifelse{and the \cl\ toolset}{} is that it should justify
\emph{indefeasible} confidence in its top claim, meaning that in
addition to confidence that the claim is \texttt{true}, we must also
be confident that there are no overlooked or unresolved doubts that
could change that judgement
\cite{Rushby:Shonan16,Bloomfield&Rushby:confidence21}.  We refer to
any concern about a case as a \emph{doubt} and we annotate the case by
adding a doubt node to the graphical representation of the assurance
argument, pointing to a node that is under suspicion.  The doubt node
contains a claim indicating the nature of the doubt (e.g., ``I think
there is something wrong here'').  At some point, we must return to
investigate the nature and origin of the doubt and will either dismiss
it as unwarranted, or refine and sharpen it into a \emph{defeater}
with a possibly more specific (counter-)claim (e.g., ``the
justification for this step is inadequate'') whose investigation is
recorded in a subcase attached to the defeater.  Thus, a doubt is
simply a defeater that has not yet been investigated (i.e., has no
subcase) and so we will generally refer to both as
defeaters.  The back and
forth investigation of an assurance case argument against doubts and
defeaters is an application of the \emph{Socratic} or
\emph{dialectical} methods for exposing error and refining
beliefs.\footnote{In philosophy, the Socratic method is considered an
instance of dialectic \cite{dialectic-wiki}; the precise distinctions
do not concern us here.}  These date back to ancient Greece but retain
their potency.  In particular, defeaters play a dialectical r{\^o}le
in argument that is similar to falsification in science
\cite{Lakatos}.  Thus, identification of potential defeaters should
not be seen as criticism but as a contribution to the development and
clear formulation of an assurance case and part of a process to
establish its indefeasibility.  In addition, developers should
consciously generate doubts, and vigorously investigate their
associated defeaters as a guard against confirmation bias, and
evaluators may raise potential defeaters as a way to elicit additional
explanation or to clarify their understanding of some part of an
assurance case.

This report is concerned with the representation, evaluation, and
recording of defeaters within an assurance case.  It does not address
systematic search for defeaters, which is an important topic akin to
hazard analysis in systems (a defeater for an argument is like a
hazard to a system) and, indeed, some defeaters may reveal previously
unconsidered hazards.  We will examine methods of searching for
potential defeaters in a separate document.

Because they make (contrary) claims, the \clasce\ graphical tool
\ifelse{}{for \atwo} uses the same oval node for defeaters as for
claims, but they are colored differently (red for
defeaters).  An example (described in more detail in
Section \ref{leds}) is shown in Figure \ref{expldef}, where a defeater
(at upper right) attached to the sideclaim for the decomposition block
claims that there are overlooked cases in the decomposition.  This
defeater has no subcase below it, so it has not yet been investigated
and represents a doubt.  The presence of a doubt or incompletely
investigated defeater causes the node that it indicates to be
considered \texttt{unsupported} and this will propagate (as described
in Section \ref{prop}) to the top claim and thereby prevents the
assurance case being considered complete or ``closed.''  Investigation
of a doubt/defeater will be recorded as an assurance subcase that
confirms or refutes its claim, as shown (for a more developed
treatment of the same example) in Figure \ref{twodef}.  

\begin{figure}[t]
\begin{center}
\includegraphics[width=\textwidth]{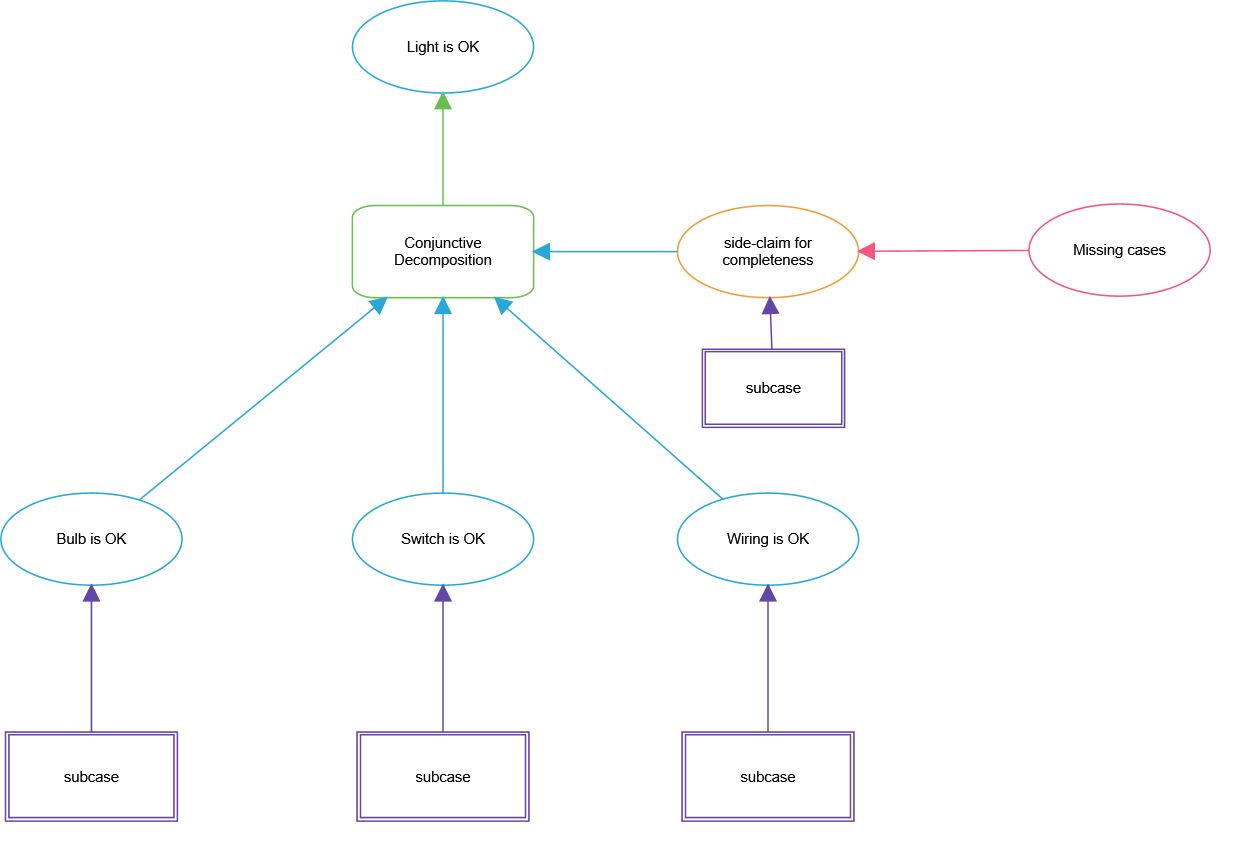}
\end{center}
\vspace*{-3ex}
\caption[Doubt as a Defeater with No Subcase]{\label{expldef}Doubt as a
Defeater with No Subcase (upper right)}
\end{figure}

If a defeater is supported by an assurance subcase that is adjudged to
be sound, so that its claim is \texttt{true}, then the defeater is
said to be confirmed or \emph{sustained} and the original case, and
possibly the system it is about, must be modified to overcome the flaw
that has been identified.  (Alternatively, the flaw may be explicitly
accepted as a \emph{residual risk}, provided it is judged suitably
insignificant \cite[Section
6]{Bloomfield&Rushby:confidence21}.)\footnotet{\clasce\ does not yet
provide methodological support for residual risks.}

After these modifications, the defeater and its subcase will (or
should) no longer apply, but we might like to retain them in the case
as documentation to assist future developers and evaluators.  Thus,
\clasce\ allows a defeater to be marked as \emph{addressed} and it and
its subcase are then treated as a comment.  Because the defeater does
not apply to the modified primary case, a narrative description of the
original problem and its resolution should be added to the defeater
node.  This may be difficult to understand (because the context is
changed in the modified case), so an alternative is to modify the
previously sustaining subcase for the defeater into a refuted subcase
(see below) for the now modified primary case.

If we suspect that a defeater is a ``false alarm,'' or it is one that
has been overcome by modifications to the original case (as above),
then our task is to \emph{refute} it: that is, to provide it with a
subcase that shows it to be \texttt{false}.  One way to do this is
with a second-level defeater that targets the first defeater or some
part of its subcase (for example, the assurance case shown in Figure
\ref{twodef} has a second defeater that attacks one of the claims in
the subcase for the first).  If the assurance subcase for that
second-level defeater is adjudged to be good, then the first defeater
is said to be \emph{refuted} and it and its subcase play no part in
the interpretation of the primary case, but can be retained as a kind
of commentary to assist future developers and evaluators who may
entertain doubts similar to that which motivated the original
defeater.

Another way to initiate refutation is by means of
\emph{counter-evidence}: that is, evidence that contradicts the claim
it is meant to support (for example, test evidence may reveal
failures).  We discuss refutational reasoning in general in Section
\ref{ref} below.

\subsection{Representation of Defeaters}

We first recap our terminology for the elements of graphical assurance
cases in \clasce.  An assurance case is composed of various kinds of
\emph{nodes}, each having a distinctive shape (e.g., oval for claims,
and rounded rectangle for argument nodes).  An argument (building) \emph{block}
consists of an argument node with a parent claim (usually drawn above
it), optional sideclaims (usually drawn to the side), and subclaims or
evidence (usually drawn below it).  The parent claim of one block will
be a subclaim or sideclaim to another block (except for the top
claim).  Thus, the overall argument forms a tree (sometimes with
cross links, so it is technically a graph).  There are just five kinds
of argument block: concretion, substitution, decomposition,
calculation, and evidence incorporation.

A consequence of our determination that defeaters function more like a
critique or commentary than part of the logical evaluation of an
assurance case is that in \cla\ we allow defeaters to be attached
(i.e., point) to any node in an assurance argument.  Specifically, in
the graphical representation of arguments used in \clasce, a defeater
node has the same oval shape as a claim node (but is colored
differently---red instead of blue) and contains a logical claim, but
we allow it to point to another claim with no intervening (rounded
rectangular) argument node.  Similarly, it may point to an argument,
evidence, or subcase node without being a subclaim or sideclaim of
that node.  And, of course, it can point directly to another defeater
node.  To aid visual recognition of defeaters and their subcases, we
recommend that defeater nodes are placed to the side of (rather than
below) the node they point to and that their subcase is developed
below them.  This allows defeater subcases to be isolated and
highlighted, as portrayed in Figure \ref{highlight}.

\begin{figure}[ht]
\vspace*{-1.1in}
\begin{center}
\hspace*{-1.0in}\includegraphics[width=8.0in]{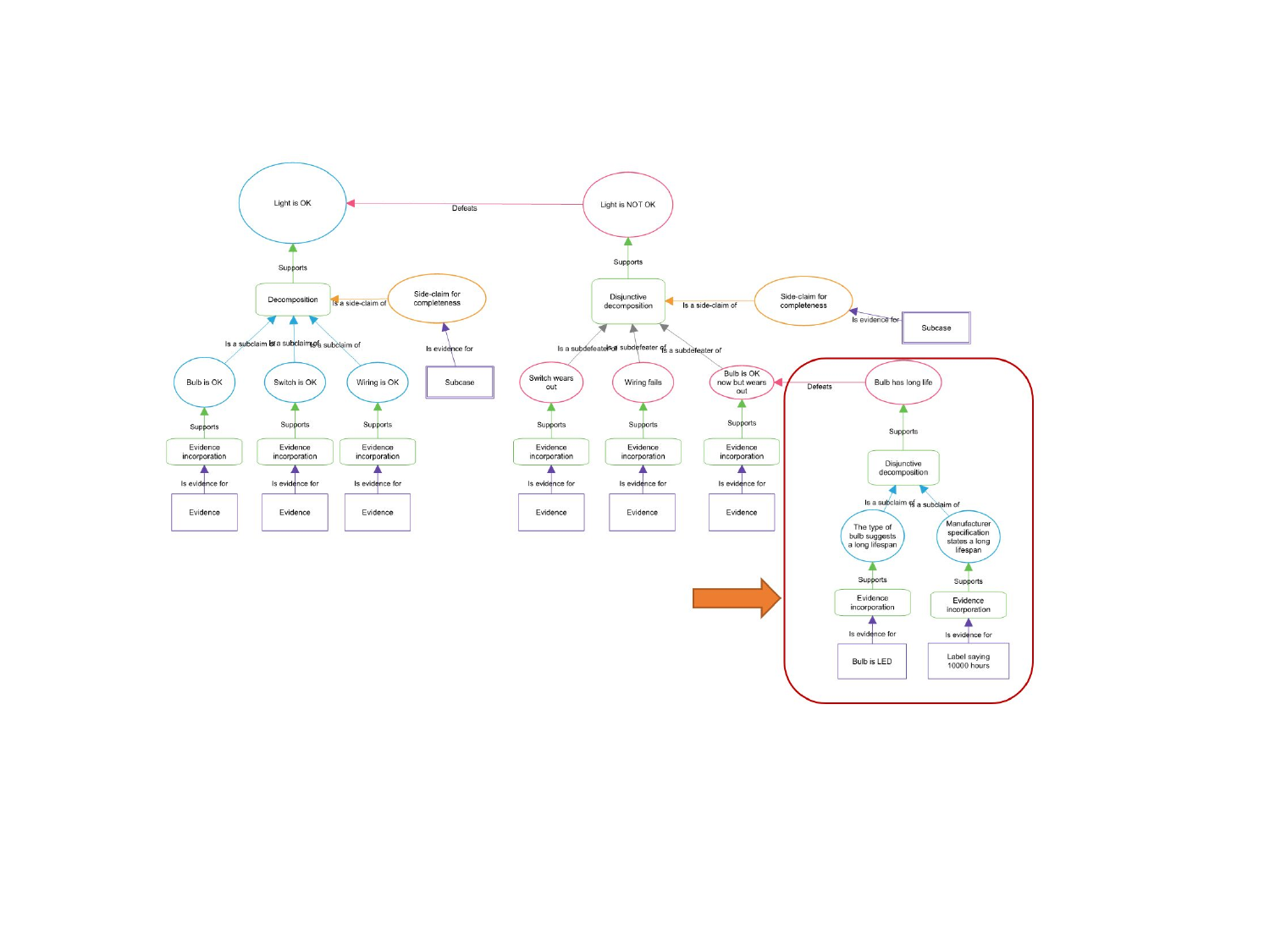}
\end{center}
\vspace*{-3ex}
\vspace*{-1.5in}
\caption{\label{highlight}Highlighted Defeater Subcase}
\end{figure}

The claim in a defeater may be vague (e.g., ``I think there's
something missing here''), which is appropriate when the defeater is
indicating an as yet unexplored doubt, or it may be specific, such as
``this claim is false,'' where ``this claim'' is the one pointed to,
or ``the justification in this argument node is inadequate.''  It is a
human judgment whether the claim in the defeater truly affects the
credibility or relevance of the node pointed to, but in the
propagation of suspicion performed by \clasce\ (described in Section
\ref{alg}) it is assumed that defeaters, if \texttt{true}, really do
``defeat'' the node pointed to.

To deal systematically with refutation and with defeaters at several
levels, we need to extend our tool support to include
\emph{refutational reasoning}.  That is to say, \clasce\ needs to
assess whether claims may be \texttt{false}, in addition to its prior
assessment of whether they are \texttt{true} or \texttt{unsupported}.
We develop rules for determination and propagation of these
assessments in the following sections.

\section{Assessing Arguments With Defeaters}
\label{prop}

We saw that investigation of a refuted defeater could introduce a
second-level defeater and, in general, investigation and resolution of
defeaters may lead to defeaters at multiple levels.  Thus, the focus
of this report is how to assess assurance cases in the presence of
defeaters, including those at multiple levels.  Since the argument for
an assurance case in Assurance 2.0 employs Natural Language
Deductivism (NLD) \cite[Section 1.3]{Bloomfield&Rushby:confidence21},
it represents something close to a logical proof,\footnote{It is close
rather than equivalent to a logical proof because an assurance case
may choose to accept residual doubts, such as nondeductive reasoning
steps, or imperfectly justified premises.}
\setcounter{ofootnote}{\value{footnote}} and it might seem that we
could look towards logical notions such as nonmonotonic logics and
defeasible reasoning\footnote{These topics are outlined in our
Confidence Report \cite[Section 5.1]{Bloomfield&Rushby:confidence21}.}
for ideas on how arguments with defeaters should be interpreted.

However, the goal of nonmonotonic logic and of defeasible reasoning is
to work out what can be concluded when there are contradictory
premises or when exceptions are added to premises (e.g., a premise
``birds fly'' gains the exception ``unless they are penguins''), but
these concerns do not apply in the same way to assurance cases.  In
\cla, defeaters are a transient phenomenon and are ``active'' (as
opposed to serving as commentary) only during development or
exploratory assessment of the case.  So, when we say that our topic is
how to assess assurance cases in the presence of multiple levels of
defeater, we do not mean how to evaluate the truth of contested claims
but how to determine which parts of a case \emph{are} contested---that
is, called into question by active defeaters, and therefore considered
``unclosed.''

Because we do not fully evaluate assurance cases with active
defeaters, but merely assess which parts of the case are closed and
which parts are still open to question, the actual claim made by a
defeater plays little part in the analysis.  In particular, if a
defeater with claim $X$ pointing to a claim $A$ is sustained, we do
not suppose that some logical combination of $A$ and $X$ is thereby
justified; we accept that the claim $A$ is challenged and revise it
and/or its supporting subcase to overcome the source of doubt.  Of
course, as noted earlier, we must make the human judgement that $X$
has some impact on the credibility or relevance of $A$ but we do not
reduce this to some logical requirement such as $X \equiv \neg
A$.\footnote{We use and $\neg$ for negation, $\wedge$ for conjunction,
$\vee$ for disjunction, $\supset$ for material implication, and
$\equiv$ for equivalence.}  Having said that, in Section \ref{elim} we
will introduce a circumstance where we do recognize the special case
where the claim in a defeater is the negation of that in the node that
it points to; we call these \emph{exact} defeaters (and the general
kind are then known as \emph{exploratory} defeaters).

\label{ref}

When a first-level defeater is sustained, the assurance case is
unsound and must be adjusted; furthermore, the system concerned may be
flawed or even unsafe and can require adjustment also.\footnote{When a
defeater reveals a flaw in the system, one possible response is to add
a runtime monitor \cite{Rushby:RV08,Littlewood&Rushby:TSE12} that will
detect and mask the hazardous condition; a systematic approach can be
developed along these lines \cite{Hawkins&Conmy:safecomp23}.}  Thus,
sustained first-level defeaters are a serious matter and should be
transient phenomena that arise only briefly during development or
while exploring a case during assessment, and are then fixed.  Hence,
the assurance goal most generally associated with first-level
defeaters is to refute them; this annuls the doubt that motivated the
defeater and leaves the original case unscathed.

One way to refute a first-level defeater is to add a second-level
defeater, but this must not degenerate into a cascade where a claim
$A$ is challenged by a first-level defeater with the counter-claim
$\neg A$ and a second-level defeater challenges that with the
counter-counter-claim $\neg\,\neg A$; this is equivalent to $A$ in
classical logic and we have achieved
nothing.\footnote{\label{class-intuit}The reference to \emph{classical
logic} is in opposition to \emph{intuitionistic logic}, which eschews
the law of the excluded middle or, equivalently, the law for
elimination of double negation.  We briefly discuss this topic in our
Confidence Report \cite[page 57]{Bloomfield&Rushby:confidence21}.}
However, the claim in a defeater need not be the negation of the claim
that it points to (and, indeed, it can point to a node that is not a
claim); it merely needs to assert something that calls the targeted
claim or node into question.  Thus, a cascade of defeaters may be
acceptable when they are based on different concerns.  Usually,
however, lower-level defeaters (i.e., those that challenge other
defeaters) should dispute claims in the \emph{subcase} of the defeater
above them: thus, the subcase for a defeater claiming $A$ may
decompose into subclaims for $B$, $C$, and $D$ and it is reasonable
for a second-level defeater to challenge $B$ (Figure \ref{twodef}
illustrates this).

As noted earlier, dealing systematically with refutation and with
defeaters at several levels, requires our tool support to be extended
with \emph{refutational reasoning}.  That is to say, \clasce\ needs to
assess whether claims may be \texttt{false}, in addition to its prior
assessment of whether they are \texttt{true} or \texttt{unsupported}.
We develop rules for determination and propagation of these
assessments in the following subsections, starting with positive
assessments, and then proceeding to refutations.

\subsection{Propagation of Positive Assessments}

Prior to the introduction of defeaters, \clasce\ needed to consider
only positive arguments, and did so using Natural Language
Deductivism, NLD\@.  That is, assessment focused on the credibility
and ultimately the indefeasibility of each argument block.  In
particular, developers and assessors would assure themselves that the
parent claim to each reasoning block was indefeasibly entailed by its
subclaim(s), given any side claim, and that the narrative
justification provided adequate documentation for this.  When evidence
incorporation blocks also passed equivalent scrutiny, and assumptions
likewise, then it could be concluded by compositionality that the
overall assurance case provided indefeasible support for the top
claim.

More formally, the argument blocks of an assurance case argument
function as logical premises with the top claim as conclusion.  The
case is then logically \emph{valid} if the argument blocks ``fit
together'': that is, the claims of the argument should form a tree
where the subclaims and sideclaim of each argument block match the
parent claim of another block, with the exception of claims
representing assumptions and residual risks (the rule is simple
because an assurance argument uses only propositional logic).  This is
enforced automatically by the graphical construction of arguments used
in \clasce: the \emph{same} claim node is used for those that should
match, connected by explicit arrows to the argument nodes concerned.

A logically valid argument is (logically) \emph{sound} if we believe
its premises to be \texttt{true}: for \clasce, this means the
narrative provided for each argument block must be judged to provide
(indefeasibly) credible justification for the argument step
represented by that block.  Notice that whereas validity concerns only
the ``form'' of the argument (i.e., do the claims match), soundness
concerns the \emph{meaning} of the claims and argument blocks involved.

\clasce\ could record assessment of soundness by attaching a checkmark
to each argument or evidence incorporation node, so that developers or
assessors can check the box if they consider the node's narrative does
provide indefeasible justification for its parent claim, and not
otherwise.  However, \clasce\ does not do this: instead, it makes the
default assumption that argument and evidence incorporation nodes are
sound, and defeaters are used to indicate otherwise.  We prefer to use
defeaters rather than checkmarks for this purpose because the defeater
and its subcase can explicitly record the reason for unsoundness.

This use of defeaters means that assessment of soundness is now
represented in the argument and thereby enters into determination of
validity: for example, if we point a defeater to an argument node and
claim ``I do not believe this narrative justification,'' then
(un)soundness of that argument node is represented in the logical
evaluation of the defeater, and hence in validity of the overall case.
This is useful because it eliminates the need for a separate check for
soundness and for ways to perform and indicate its propagation: it is
all handled by the mechanisms for validity and defeaters.

However, determination of validity becomes a little more complicated
than before because we have to account for defeaters, and we also
allow incompletely developed arguments.  In these cases, we cannot
expect the top claim to be adjudged \texttt{true}; instead, we can ask
which claims \emph{are} \texttt{true} and which are
\texttt{unsupported} (i.e., have contested or incomplete subcases),
and we can do this by propagating logical assessments upward from the
leaf nodes.\footnote{\label{up-down}It can be argued that doubt should
propagate downward as well as upward: for example, if a concretion
block below the top node is challenged, then surely it calls the whole
development into question as it suggests that the case is establishing
an incorrect claim.  Unfortunately, determination of which parts of a
case should be adjusted to correct a contested argument cannot be
reduced to calculation.  Hence, there can be no uncontested rule for
propagation of doubt.  The purpose of propagation is, firstly, to
determine if an argument is valid and, secondly, to direct attention
to those parts most likely in need of attention if it is not, and we
consider that our approach accomplishes this in a manner that is
effective and readily understood.}

There are five kinds of leaf nodes: claims lacking a subcase
(typically indicating undeveloped parts of the case), which are
considered \texttt{unsupported}; assumptions, which are claims lacking
a subcase that are specially designated and justified, and are
assessed \texttt{true}; residual risks, which are are defeaters
lacking a subcase that are specially designated and justified, and are
assessed \texttt{false}; references to external subcases and theories,
which inherit any existing assessment for the top claim of the subcase
or theory instantiation concerned, and are otherwise assessed as
\texttt{unsupported}; and evidence nodes, which are rather more
complicated and are discussed later.

The parent claims of interior argument blocks (i.e., concretion,
substitution, decomposition, and calculation) are assessed
\texttt{true} when all their subclaim(s) and their sideclaim (if any)
are assessed \texttt{true}; otherwise they are \texttt{unsupported}.
As explained above, the narrative justification provided with each
argument node plays no part in the propagation of these assessments,
which are concerned only with logical \emph{validity}.  The narrative
justifications are critical in assessing \emph{soundness} and
developers are expected to ensure (and evaluators to confirm) that the
narrative supplied for the argument node of each block justifies
(indefeasible) belief in the parent claim, given its subclaims and any
sideclaims.  During development or, later, evaluation, any concern about
the soundness of a narrative justification can be registered by
attaching a defeater to the argument node concerned, which will then
be factored into determination of validity.

Notice that these interior argument blocks represent premises that are
\emph{a priori}, meaning that we believe them by virtue of thinking
about, and understanding, the system and the claims and the argument
block concerned.  This is in contrast to evidence incorporation blocks,
which are \emph{a posteriori} premises, meaning that our belief about
them rests on observations or measurements of the system and its
environment.

We now turn to assessment of evidence incorporation blocks.  As with
other argument blocks, for the purposes of logical validity we could
just check that the evidence is available and assume that it is good
(and will be challenged by a defeater if not) and assess the parent
claim as \texttt{true}.  However, evidence incorporation blocks are
different to other blocks in that they are not solely focused on
logical reasoning but provide the bridge between logical reasoning and
the external world---which is manifested as evidence.  Hence, we must
choose how much of the evaluation of evidence we wish to factor into
determination of logical validity, and how much into soundness.  The
choice made in \clasce\ is that validity considers only the
\emph{presence} of the evidence, not its merit; interpretation and
evaluation of evidence is the responsibility of argument blocks above
the evidence incorporation block.

Specifically, the parent node of an evidence incorporation block is a
claim that usually states ``something measured'' about the evidence
(e.g., ``the tests achieve MC/DC coverage''): that is, it indicates
what the evidence \emph{is}.  Above that there is usually a
substitution block whose parent claim states ``something useful''
derived from the evidence (e.g., ``there is no unreachable code''):
that is, it indicates what the evidence \emph{means} (see
\cite[Sections 1.3 and 2.2]{Bloomfield&Rushby:confidence21}).  In the
interests of grammar, we will call these \emph{evidentially measured}
and \emph{evidentially useful} claims, respectively.\footnote{Here, we
are assuming the evidentially useful claim is supported by an argument
block directly above the evidentially measured claim.  It is possible
that additional reasoning is required (e.g., adequacy of tools
involved in generating the evidence) so that the useful claim is
further removed from the measured claim; alternatively the additional
reasoning could appear in a sideclaim to the substitution block
relating the two evidential claims.  Further practical experience is
needed to develop recommendations for this topic.}

The narrative justification of the evidence incorporation node is
expected to support (indefeasibly) the determination that the supplied
evidence really does deliver the measured results (given any
sideclaims).\footnotet{\clasce\ does not support sideclaims
to evidence incorporation blocks; this is to prevent the argument
continuing indefinitely with evidence requiring more evidence.  However, evidential sideclaims can
ground out in assumptions or residual risks so this may be revisited
in future.}  If the narrative justification for this is considered
inadequate during development or evaluation, then a defeater can be
attached to the evidence incorporation node.  As with other argument
blocks, the narrative justification plays no part in assessing the
logical validity of an evidence incorporation block; thus, its
evidentially measured claim is assessed \texttt{true} provided the
evidence is present; otherwise it is
\texttt{unsupported}.\footnotet{\label{present}Currently, When an evidence
incorporation block is added to a \clasce\ assurance argument, the
evidence is required to be present or the block will not be added.
However, the evidence could subsequently fail (e.g., if it requires
reference to an external data source).  Currently, \clasce\ does not
monitor this; hence proposals for an evidence \texttt{present} flag.}

Measured evidence is transformed into useful evidence by a
substitution block placed above the measured claim of the evidence
incorporation block.  Usually, we expect the measured evidence to
affirm the evidentially useful claim and we use \emph{confirmation
measures} to help make this determination in a principled way
(confirmation measures are discussed in detail in our Confidence
Report \cite[Section 2.2]{Bloomfield&Rushby:confidence21}).  An
example is Good's confirmation measure $\log\frac{P(E \vbar C)}{P(E
\vbar \neg\, C)}$, where $E$ represents the evidence and $C$ the
evidentially useful claim.  It is not necessary to apply these
measures numerically: what matters are the concepts underlying them,
so that in Good's measure we are asked to consider the likelihood of
the evidence given the claim \emph{vs}.\ its likelihood given the
counterclaim.  This is intended to ensure that the evidence not only
supports the claim but that it discriminates between this claim and
others (and the counterclaim in particular).  Developers and
evaluators can use these measures informally or can apply them to
numerical or qualitative (e.g., low, medium, high) estimates of the
subjective probabilities involved, if they find it useful to do so.
Conceptually or numerically large positive confirmation measures
indicate highly affirming evidence and this will be recorded and
explained in the narrative justification of the substitution block,
which can be challenged by a defeater in case of doubt.\footnotet{The
current \clasce\ has an experimental plugin for confirmation measures,
but does not provide methodological support for their use in
assessment of evidence.}  The usual rule for logical validity of a
substitution block applies, so the parent (i.e., evidentially useful)
claim will be assessed \texttt{true} provided the (evidentially
measured) subclaim and any sideclaims are also \texttt{true}.  Also as
usual, the narrative justification, supported by a confirmation
measure, is assumed to affirm soundness of this determination, and
will be challenged by a defeater otherwise.  

However, the usual calculations can be overridden---because evidence
sometimes refutes a claim (e.g., when tests reveal a failure), which
transforms it into \emph{counter-evidence}.  This is examined below,
where we consider propagation of assessments in refutational
arguments.

\subsection{Propagation With Refutations}
\label{alg}

\textbf{Note}: the \clasce\ implementation does not currently support
refutational reasoning.

In refutational arguments, we need to consider the possibility that
claims may be \texttt{false}, in addition to \texttt{true} or
\texttt{unsupported} as considered above.\footnote{There is a subtle
point concerning the interpretation of negative assessments, where
\texttt{unsupported} can sometimes function like \texttt{false}.  If a
top claim assessed as \texttt{unsupported} states something like ``the
system is safe'' then more work is needed to refine the case so that
it delivers a definitive assessment (i.e., \texttt{true} or
\texttt{false}).  But if the claim is ``\emph{the argument establishes
that} the system is safe'' then \texttt{unsupported} carries stronger
significance and its external interpretation will be the same as
\texttt{false}.}  There are two ways that \texttt{false} assessments
may be introduced into an argument: one is through counter-evidence,
and the other is via defeaters; we begin with counter-evidence.

The discussion here is a continuation of that in the previous section
and concerns the case where evidence does not merely fail to affirm
its evidentially useful claim, but contradicts it.  We considered
affirming evidence in the previous section: this is evidence that
delivers a strong positive confirmation measure and justifies the
assessment that its evidentially useful claim is \texttt{true}; we now
consider the other cases.  Conceptual or numerical confirmation
measures close to zero indicate weak evidence and in this case the
evidentially useful claim will be assessed \texttt{unsupported}.
But sometimes evidence contradicts the claim it is meant to support:
for example, as noted previously, testing may reveal failures.  We
refer to this as \emph{counter-evidence} and it should lead to
a strongly negative confirmation measure.  In this case, we assess the
evidentially useful claim as \texttt{false}, provided its evidentially
measured subclaim and any sideclaims are \texttt{true}; otherwise it
is \texttt{unsupported}.

We next turn to analysis of defeaters.  Since a defeater can point to
any kind of node, we define the claim \emph{affected by} the defeater
to be the node pointed to if this is a claim or defeater, and
otherwise the parent claim (which may be a defeater) of the node
pointed to.

When the claim in a (non-exact) defeater is assessed \texttt{false} it
means the defeater is refuted; hence, the main case (or subcase for
lower-level defeaters) is exonerated and its claims are assessed as if
the defeater were absent.  Exact defeaters (those that point directly
to a claim or other defeater and whose claim is the negation of that
pointed to) are a special case that is considered later, in Subsection
\ref{exact}.

When the claim in a defeater is assessed \texttt{unsupported} (which
also applies when the defeater has no subcase---i.e., it is merely a
doubt), then so is the claim affected by the defeater.  And when the
claim in the defeater is assessed \texttt{true}, then the affected
claim is also assessed \texttt{unsupported};\footnote{It cannot be
assessed \texttt{false} because the defeater may not precisely refute
the affected claim (unless it is an exact defeater, which is
considered later), but merely call it into question.}  again, exact
defeaters are a special case and will be considered later.

These assessments override the assessments due to any other nodes
pointing to the affected claim (which may affirm it as \texttt{true}:
when a claim is challenged by a \texttt{true} or \texttt{unsupported}
defeater, we have to accept that it is unresolved.  And this is true
even if the defeater has no subcase (i.e., is a doubt); the mere
existence of the doubt calls the affected claim into question.  These
assessments are also independent of the logical relationship between
the affected claim and the claim in the defeater that challenges it
(again, excluding exact defeaters): we have made a human judgement
that the defeater does call the node pointed to (and hence the
affected claim) into question, even if thee two claims are logically
unrelated.  As a result, there is some ambiguity here: when the claim
affected by a defeater is assessed as \texttt{unsupported}, it could
either be because the defeater's subcase has sustained the defeater or
because that subcase is incomplete; we need to examine the assessment
of the defeater (\texttt{true} or \texttt{unsupported}, respectively)
to discriminate the two cases.  In the latter case, the defeater's
subcase needs more work, while in the former the main argument needs
to be revised (and possibly also the system concerned).

Finally (apart from exact defeaters), we consider propagation of
\texttt{false} assessments through the remaining kinds of argument
blocks (i.e., concretion, substitution, decomposition, and
calculation).  In NLD, individual argument blocks of these kinds are
intended to be deductively valid: that is, they are interpreted as
material implications of the form
\begin{equation}
\label{deduction}
\texttt{sideclaim} \wedge \texttt{subclaims} \supset \texttt{parent claim}
\end{equation}
where the subclaims, if there is more than one, are usually conjoined
(we introduce disjunctive subclaims in Section \ref{conjdisj}).

The sideclaim and subclaims constitute the \emph{antecedent} to this
implication.  As described in the previous section, when all claims in
the antecedent are assessed \texttt{true} then, by the rules of
classical logic, so is the parent claim.  And if any antecedent claims
are \texttt{unsupported}, then the parent claim is also.  But suppose
some claims in the antecedent are assessed \texttt{false}.  Since they
are conjoined, this means the whole antecedent is \texttt{false}; does
this mean we should assess the parent claim as \texttt{false} too?

It does not: it would be attempting to derive $\neg A \supset \neg B$
from $A \supset B$, and this is the logical fallacy of ``denying the
antecedent'' \cite{denying-antecedent-wiki}.\footnote{An informal
illustration of denying the antecedent uses the subclaim/premise ``if
college admission is fair, then affirmative action is unnecessary'' to
fallaciously infer the claim/conclusion ``college admission is not
fair, so affirmative action is needed.''}  Moreover, there is a
further problem: if the antecedent is \texttt{false}, then it can
imply anything: this is the \emph{false implies everything}
problem.\footnote{$A \supset B$ is equivalent to (or is defined as)
$\neg A \vee B$ so, if $A$ is \texttt{false}, $\neg A$ is
\texttt{true}, and $\neg A \vee B$ is \texttt{true} independently of
$B$.}  Thus, in general, we cannot propagate \texttt{false} upward
through these four kinds of assurance blocks;\footnote{There is a
special case where the (conjunction of) subclaims is \emph{equivalent}
to the parent claim (given the sideclaim) rather than merely entailing
it: it is legitimate to propagate \texttt{false} in this case but
\clasce\ does not do so (because it does not attempt to interpret the
language of claims).  Instead, the case should be modified to use an
exact defeater.} we must do something weaker and the appropriate
response is to assess the parent claim as \texttt{unsupported}.

\subsection{Exact Defeaters}
\label{exact}

As noted above, exact defeaters are a special case; their purpose is
to introduce negation into an assurance case and this primarily finds
application in an alternative form of argumentation to be described in
Section \ref{elim}.

An exact defeater is one that: a) points to a node that is either a
claim or another defeater that b) lacks a subcase, and c) whose
own claim is the negation of the one pointed to.  An example is shown
in Figure \ref{elimdef}; the example in Figure \ref{expldef} cannot be
an exact defeater, independently of its claim, because the claim it
points to has a subcase, contradicting b) above.

Because claims in \clasce\ are written in natural language, it is not
trivial to determine if one claim is the negation of another.
Accordingly, \clasce\ provides an explicit selection in its interface
to indicate that a defeater should be treated as the exact negation of
the claim or defeater that it points to.  Furthermore, the node
pointed to may have a subcase, but it will be ignored (and indicated
so in the graphical presentation) when the node becomes the target of
an exact defeater.This is to support exploratory development of a
case without having to undo or redo previous work.

The propagation rules for exact defeaters are simple: if the exact
defeater is assessed \texttt{unsupported}, then so is the node that it
points to; otherwise the assessment of the node pointed to is the
logical negation of the assessment of the claim in the defeater.

\subsection{Summary of Propagation Rules}

Here we present a summary of the propagation rules for truth
assessments described in the previous subsections.  Remember, we are
using a three-valued logic: \texttt{true}, \texttt{false}, and
\texttt{unsupported}.

\begin{description}

\item[Assumptions:] assigned \texttt{true}

\item[Unsupported claims] (i.e., claims with no subcase):
assigned \texttt{unsupported}

\item[External subcases:] parent claim inherits whatever the subcase delivers for
  its top claim, with default assignment \texttt{unsupported}

\item[Evidence Incorporation:]
if evidence \emph{present}  \\then
  (\emph{evidentially measured})
  parent claim is assigned \texttt{true},\\ otherwise \texttt{unsupported}
(see below for \emph{evidentially useful} claims)

\item[General assurance blocks] (concretion, substitution,
    decomposition, and \\ calculation): if all subclaims and sideclaim
    \texttt{true}, \\ then parent claim is assigned \texttt{true},
 otherwise \texttt{unsupported}

\item[Special case] for substitution blocks delivering
  \emph{evidentially useful} parent claims:\\
  Justification should reference confirmation measures then, provided
  sideclaim and evidential measured subclaim are \texttt{true},\\
if \hspace*{0.1em}confirmation measure is \texttt{strongly positive}, then parent claim is \texttt{true}\\
\hspace*{1em}confirmation measure is \texttt{neutral}, then parent claim is \texttt{unsupported}\\
\hspace*{1em}confirmation measure is \texttt{strongly negative}, then parent claim is \texttt{false}\\
otherwise \texttt{unsupported}

\item[Ordinary (non-exact) defeaters] (includes doubts):
  if claim in defeater is \texttt{false},\\ then
   rest of the case is unaffected (defeater is defeated),\\
 otherwise \emph{affected claim} is \texttt{unsupported}

\item[Exact defeaters:] \mbox{}\\
if claim in defeater is \texttt{false}, then parent claim is \texttt{true}\\
\hspace*{1em}claim in defeater is \texttt{true}, then parent claim is \texttt{false}\\
\hspace*{1em}claim in defeater is \texttt{unsupported}, then parent claim is \texttt{unsupported}

\end{description}

\subsection{Logic Programming Interpretation}

Our interest in this topic is motivated by the possibility that a
suitable Logic Programming language could mechanize the propagation
rules described in the previous subsections.  Although these rules are
easy to implement directly in \clasce, translation to Logic
Programming could allow additional automated exploration and analysis
that provide added value.

As noted in formula (\ref{deduction}), \clasce\ argument blocks are
interpreted as a material implication, which can be
rewritten\footnote{Here we are replacing $A_1\wedge\cdots\wedge{}A_n
\supset B$ by $\neg(A_1\wedge\cdots\wedge{}A_n) \vee B$, applying
commutativity of $\vee$, and then using De Morgan's law to replace
$\neg (A_1 \wedge\cdots\wedge A_n)$ by $\neg A_1 \vee\cdots\vee\neg
A_n$.}  as
\begin{equation}
\label{horn}
\texttt{parent claim} \vee \neg \texttt{sideclaim} \vee \neg
\texttt{subclaim}_1 \vee \cdots \vee \neg \texttt{subclaim}_n.
\end{equation}
The claims are \emph{ground terms} (i.e., they contain no variables)
and a ground term or its negation is called a \emph{literal}; a
disjunction of literals such as (\ref{horn}) is a \emph{clause}.  A
clause with exactly one positive (i.e., unnegated) term, again such as
(\ref{horn}), is called a \emph{definite clause}; one with with no
positive terms is called a \emph{goal}; and a single positive term is a
\emph{fact}; together, these constitute \emph{Horn clauses}.  The
arguments of assurance cases can be represented as collections of Horn
clauses: evidence and assumptions will be facts, and the other
reasoning blocks will be definite clauses.  Goals arise when we pose
questions about the argument (e.g., ``is the top claim
\texttt{true}?'').

Horn clauses are also the basis of \emph{logic programming}: a
collection of Horn clauses has an interpretation in logic (namely, the
set of literals that must be \texttt{true} to satisfy all clauses in
the collection) and another, operational, one that computes this set
(the set will be empty if the clauses are \emph{unsatisfiable}).

In Logic Programming, clause (\ref{horn}) above is usually written
\begin{equation}
\label{prolog}
\texttt{parent claim :- sideclaim, }
\texttt{subclaim}_1\texttt{,\ldots, subclaim}_n.
\end{equation}
where the term on the left side of the \texttt{:-} symbol is called
the \emph{head}, and the list of terms on the right hand side is
called the \emph{body}.  An example \texttt{fact} is written as \[
\texttt{claim} \] and indicates that this literal is \texttt{true}.

There are several Logic Programming languages; examples include Prolog
and Datalog and there are further variants within these.  A
significant source of variation is the treatment of negation and how
the logic and operational interpretations are kept aligned under these
different treatments.

In assurance cases, exact defeaters require strict, or logical (i.e.,
classical) negation because a \texttt{true} defeater claim (body)
entails a \texttt{false} target claim (head) and vice-versa.  Other
defeaters are more complex; first, the affected claim will usually
have an existing assessment derived from the primary case (i.e.,
ignoring the defeater) and a defeater can \emph{override} this.  In
particular, the affected claim becomes \texttt{unsupported} if the
defeater is \texttt{true} or \texttt{unsupported}, and is left alone
if the defeater is \texttt{false}.  This means we need a way to
represent \texttt{unsupported} and its special rules in our logic
program.  Fortunately, these resemble ``unproved'' and ``weak''
negation, respectively, in certain forms of Logic Programming.  In
particular, the form of Logic Programming known as Answer Set
Programming (ASP) supports both classical or strict negation (written
as prefix \texttt{-}) and the autoepistemic (stable model semantics)
interpretation of weak negation (negation as failure), written as
prefix \texttt{NOT}.  That is, \texttt{NOT p} is \texttt{true} if $p$
is unproved, and \texttt{NOT -p} is \texttt{true} if \texttt{-p}
(i.e., the strict negation of \texttt{p}) is unproved.  Notice that
\texttt{NOT p} and \texttt{NOT -p} may both be \texttt{true} (meaning
that neither \texttt{p} nor its strict negation has been proved),
whereas if \texttt{p} is \texttt{true} then \texttt{-p} must be
\texttt{false}, and vice-versa.

\clasce\ has a \emph{Prolog Export}
plugin that uses the s(CASP) system \cite{Gupta:sCASP22} for answer
set programming to perform semantic analysis of assurance cases
\cite{Murugesan-all:GDE23}.
Our treatment of assurance cases with defeaters is interpreted in
s(CASP) using the following translation rules, which are derived
directly from those stated informally in the previous subsection.
However, in the informal rules we implicitly assumed a ``two pass''
interpretation where, in the first pass we ignored ordinary
(non-exact) defeaters while propagating truth values and then applied
the effects of defeaters in the second pass.  Here, we will translate
the assurance argument together with its defeaters and submit it all
to the s(CASP) engine, which will interpret it as a complete program.
So we augment each rule with a \texttt{defeater} term that will be
instantiated by the translation of the claim(s) in the actual
defeater(s) pointing to this claim or its argument node (and be absent
if there are none).

\begin{description}

\item[Assumptions:] we simply state the \texttt{claim} as a fact,
provided it is not defeated.
\begin{alltt}
  claim :- -defeater
\end{alltt}

It is worth examining the cases here.  If the \texttt{defeater} is
\texttt{false} or absent, the \texttt{claim} is \texttt{true}.  If the
\texttt{defeater} is \texttt{true}, then \texttt{-defeater} is
\texttt{false}, but this does not make \texttt{claim} \texttt{false}
(the body affects the head only when it is \texttt{true}).  Finally,
if the \texttt{defeater} is unproved, then so is the \texttt{claim}.

\item[Unsupported claims] (i.e., claims with no subcase): we say
nothing about the claim; s(CASP) will treat it as \texttt{unproved}.

\item[External subcases:] the parent claim inherits whatever the
subcase delivers for its top claim.\footnotet{\clasce\ does not
currently provide a construct to reference an external subcase; it is
plausible that such a construct could be the target of defeaters,
claiming that the subcase is wrong/irrelevant etc., so this rule may
be augmented in future.}

\item[Evidence Incorporation:] \mbox{}
\begin{alltt}
  measured_parent_claim :- evidence_present, -defeater
\end{alltt}
Here, the
\texttt{evidence\_present} flag is set \texttt{true} (i.e., stated as
a fact) when the evidence is present and is not mentioned otherwise.

\item[General assurance blocks] (concretion, substitution,
    decomposition, and \\ calculation): 
\begin{alltt}
  parent_claim :- subclaims, sideclaims, -defeater
\end{alltt}
Here, \texttt{subclaims} is a list of one or more claims and
\texttt{sideclaims} is also a list of zero or more.

\item[Special case] for substitution blocks delivering
  \emph{evidentially useful} parent claims:
  Justification should reference confirmation measures, then
  \texttt{confirmation} can be set \texttt{true}, \texttt{unproved},
  \texttt{false} (i.e., respectively stated as a fact, not mentioned, stated as a
  negated fact) according to whether the confirmation measure
  is strongly positive, neutral, or strongly negative.

If confirmation measure is strongly positive:
\begin{alltt}
  useful_parent_claim :- sideclaims, measured_claim, -defeater  
\end{alltt}

If confirmation measure is strongly negative:
\begin{alltt}
 -useful_parent_claim :- sideclaims, measured_claim, -defeater  
\end{alltt}
Note that this causes the evidentially useful parent claim to be set
\texttt{false} when the evidence is strongly negative.

\item[Ordinary (non-exact) defeaters] (includes doubts):

  if the defeater is unsupported (i.e., is a doubt), then it is not
  mentioned and will default to \texttt{unproved}.  Otherwise it must
  be the parent node of some argument node and will be set according
  to its type using the rules above.

\item[Exact defeaters:] \mbox{}\\
\begin{alltt}
  parent_claim :- -defeater_claim
 -parent_claim :- defeater_claim
\end{alltt}

Note that we need two rules here: one to propagate \texttt{true} and
the other to propagate \texttt{false}.

\end{description}

\subsection{Comparison with Dialectics in GSN}

The Goal Structuring Notation (GSN) has a ``Dialectic Extension''
\cite[Section 1:6]{GSN:community-std3} that serves purposes and
provides capabilities very similar to our defeaters.  Whereas we have
nodes explicitly marked as defeaters that can point to other nodes, in
GSN a ``dialectic challenge'' is indicated by a special kind of link
(i.e., arrow) that can point to other links as well as to nodes.
(Links in GSN can indicate ``supported by'' or ``in context of''
relationships, whereas arrows in \cla\ merely identify the nodes that
constitute a block and it suffices to point defeaters at nodes.)  GSN
does not explicitly distinguish exploratory and exact defeaters, but
informally it can accomplish equivalent effects.

GSN does not make the strong distinction between logical validity and
indefeasible soundness that is a focus of \cla, nor does it perform
refutational reasoning, so argument nodes are considered \texttt{true}
or \texttt{false} (i.e., refuted), without the additional
\texttt{unsupported} valuation of \clasce.  A refuted goal or link is
indicated by a large X.  Furthermore, propagation of defeat is
performed informally and regarded as ``a matter for expert judgement
and the more likely outcome is that the original goal structure is
refactored at this point'' \cite[Section
2:11.3.4]{GSN:community-std3}.  Similarly, retention of defeaters and
``how to present dialectic argument in the final goal structure'' are
regarded as ``choices to be made by the practitioner'' \cite[Section
2:11.4/6]{GSN:community-std3}.

The dialectic extension for GSN assurance cases is supported by the
commercial version of {\sc asce} while a more complex form of
defeasible reasoning \cite[Section
5.1.4]{Bloomfield&Rushby:confidence21} is supported by the Astah GSN
tool \cite{Astah-gsn,Takai&Kido14}.  Naturally, we consider that the
treatment of defeaters in \atwo\ and their implementation in \clasce\
strike the best balance of utility and rigor.

\section{Eliminative Argumentation and Exact Defeaters}
\label{elim}

``Eliminative Induction'' is a method of reasoning that dates back to
Francis Bacon who, in 1620 \cite{Bacon-organon}, proposed it as a way
to establish a scientific theory by refuting all the reasons why it
might be false (i.e., its defeaters).\footnote{This was a precursor to
modern methods and theories of science, which see falsifiability as
the key characteristic \cite{Popper:2014logic}; however, they can be
related via Bayesian Epistemology \cite{Hawthorne93,Vineberg96}.}
Weinstock, Goodenough, and Klein \cite{Goodenough-etal:ICSE2013} build
on the idea of Eliminative Induction to develop a means of assurance
that they call \emph{Eliminative Argumentation}.  Here, instead of
attempting to confirm a positive claim such as ``the system is safe''
we instead attempt to refute the negative claim ``the system is
\emph{un}safe.'' A successful refutation will establish the negation
of that claim, namely ``the system is \emph{not} unsafe.''  In
classical logic this is equivalent to establishing the positive claim
by virtue of the rule for elimination of double negation (recall
Footnote \ref{class-intuit}), and thereby provides the desired
assurance.  Diemert and Joyce \cite{Diemert&Joyce2020} and others
\cite{Millet:CERN-LHC23} report successful application of eliminative
argumentation in assurance of real systems

\ifelse{\cl\ is based on the methodology of \atwo, which}{The
methodology of \atwo} favors \emph{positive cases} where a
constructive argument is developed in support of some beneficial claim
about a system.  Of course, in a formal or regulated safety process,
there will generally be several layers of review and challenge to the
case that should eliminate flaws and undue optimism; nonetheless, to
combat confirmation bias and other complacency, we accept that it can
be useful to explicitly consider different points of view using
contrary claims and negative arguments.  Furthermore, when attempting
to refute a defeater to a positive case, we are in a context similar
to eliminative argumentation and therefore need to support this kind
of reasoning for our own purposes.

In the framework of \cla, eliminative argumentation can be represented
by attaching an exact defeater to a positive claim, and then
attempting to refute it.  We sometimes refer to this use of defeaters
as \emph{eliminative} and to the conventional use as
\emph{exploratory}.  Notice the whereas exploratory defeaters
\emph{augment} the main argument, providing an exploratory
investigation or commentary, eliminative or exact defeaters are used
as a reasoning step \emph{within} the main argument.

\begin{figure}[t]
\begin{center}
\includegraphics[width=4.5in]{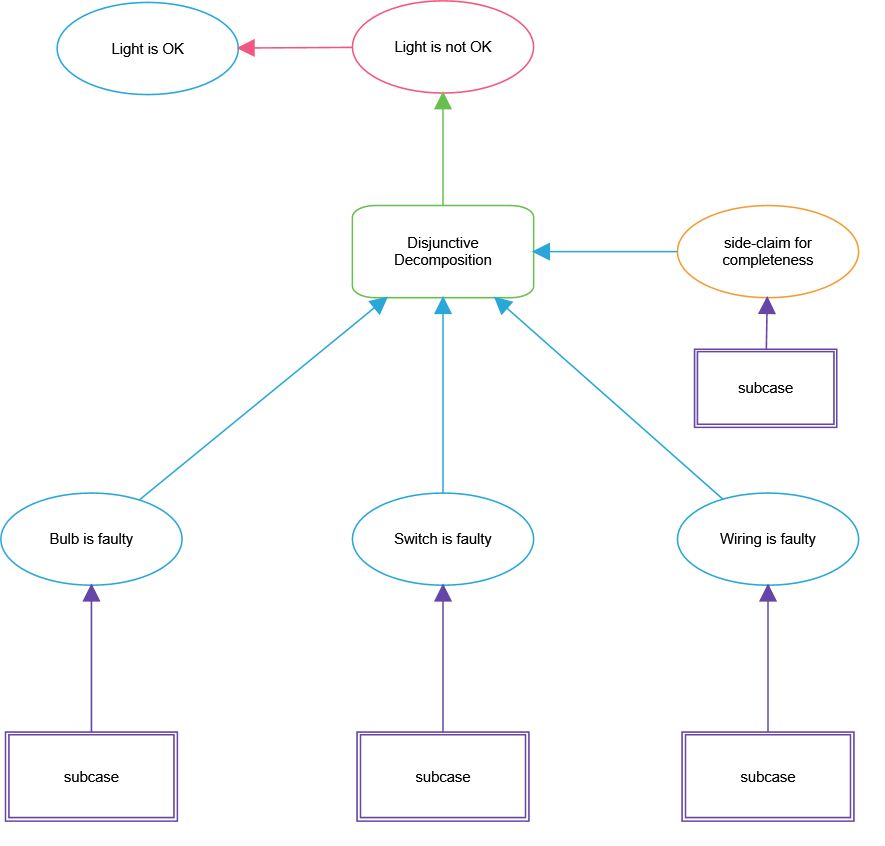}
\end{center}
\vspace*{-3ex}
\caption{\label{elimdef}Eliminative Argument Using an Exact Defeater}
\end{figure}

An eliminative argument is shown in Figure \ref{elimdef} where we
provide assurance that a newly installed electric light is OK by
introducing an exact defeater that asserts it is not OK (i.e., faulty)
and then refuting all the reasons that could make it so.  Notice that
this uses a disjunctive decomposition, which will be explained in the
following section.  See Figure \ref{twodef} for a more elaborated
development of a traditional argument for the same example, which has
a first-level exploratory defeater in the upper right and an exact
defeater (with a disjunctive decomposition) at the second-level in the
lower center.

\section{Conjunctive and Disjunctive Decomposition Blocks}
\label{conjdisj}

We stated earlier that the subclaims of decomposition blocks are
treated as a conjunction and we might wonder if this is always
appropriate.  For example, suppose we are reasoning about a fault
tolerant system $S$ that has two redundant subsystems $A$ and $B$ and
it is sufficient for safety that either one of these is working.  It
might seem that we should have a substitution block with parent claim
``$S$ is safe'' and subclaim ``$A$ is working correctly or $B$ is
working correctly,'' which would seem to invite the support of a
decomposition block where the subclaims are disjoined instead of
conjoined.  But this is not correct: it would allow us to provide a
subargument that considers only the case where $A$ is working
correctly---but we do not know in advance whether it is $A$ or $B$
that will be working correctly.  We also need to consider the case
where both are working correctly (because they might get in each
other's way).  A valid argument is a standard (conjoined)
decomposition block with three subclaims ``$A$ and $B$ are both
working correctly,'' ``$A$ alone is working correctly,'' and ``$B$
alone is working correctly.''
Another candidate for disjunctive decomposition is when we have two
(or more) alternative arguments in support of a given claim, so that
if the assessors do not like Argument 1, they can consider Argument 2.
But, again, if these are disjoined, we can establish the parent claim
by developing only Argument 1, thereby subverting the motivation for
having alternative cases.  So, once again, we should interpret
decomposition as a conjunction.

There is, however, one situation where disjunctive decomposition can
be appropriate: this is when the base assurance case is a generic
template (i.e., part of what we call a \emph{theory}) that may be
instantiated in several ways.  For example, we may have a claim $P$
that can be ensured either by subcase $Q$ or subcase $R$, depending on
the realization chosen in the actual system.  Here, we can use a
disjunctive decomposition (which needs no sideclaim) in the generic
template and elaborate only the appropriate subcase.  We give an
example in Section \ref{leds}, where the claim that a light bulb will
last a certain time can be satisfied either by using an LED, or by
using an incandescent bulb that comes with a label ``guaranteeing''
adequate life.

In our experience, and apart from their use in theories and templates,
the appropriate interpretation for the subclaims to a decomposition is
always conjunction---at least in positive cases.  But what about
negative cases: those where the local goal is to refute some higher
claim?  Here, applying eliminative argumentation to system $S$ above,
we might have the counter-claim ``$S$ is unsafe'' and we support this
with the three subclaims ``$A$ is not working safely,'' ``$B$ is not
working safely,'' and ``neither is working safely.''  But any one of
these is sufficient for $S$ to be unsafe, so this step of the argument
needs to be represented by a decomposition block in which the
subclaims are disjoined.  An explanation for this can be seen
by supposing the original, positive case can be represented as $$
A_{ok} \wedge B_{ok} \wedge AB_{ok} \supset S_{ok}.$$ Then from this
we postulate a negative case\footnote{Remember, this is invalid in
general (it is denying the antecedent), and therefore needs careful
justification; it is valid if the implication can be strengthened to
equivalence (i.e., to \emph{if and only if}).}  $$ \neg(A_{ok} \wedge
B_{ok} \wedge AB_{ok}) \supset \neg S_{ok},$$ which is equivalent (by
De Morgan's law) to
$$ \neg A_{ok} \vee \neg B_{ok} \vee \neg AB_{ok} \supset \neg
S_{ok}.$$

We conclude that in negative cases it can be useful to have a
disjunctive form of decomposition block.  One way to do this is
to maintain just a single kind of decomposition block, but to
interpret it conjunctively in positive subcases and disjunctively in
negative ones.  Alternatively, we can introduce an explicitly
disjunctive form of decomposition, and this is the approach employed
in \clasce.  We prefer
this choice because it requires an active decision by the developers
of the assurance case and explicitly records it for the assessors.
Furthermore, the disjunctive form is available for use in positive
cases (and the conjunctive form in negative cases) should this ever be
considered appropriate.

The propagation of truth values over a disjunctive decomposition
requires that if the sideclaim and \emph{any} subclaim are
\texttt{true}, then the parent claim is also \texttt{true}; otherwise
it is \texttt{unsupported}.

\section{An Illustrative Example}
\label{leds}

\begin{figure}[ht]
\begin{center}
\includegraphics[width=\textwidth]{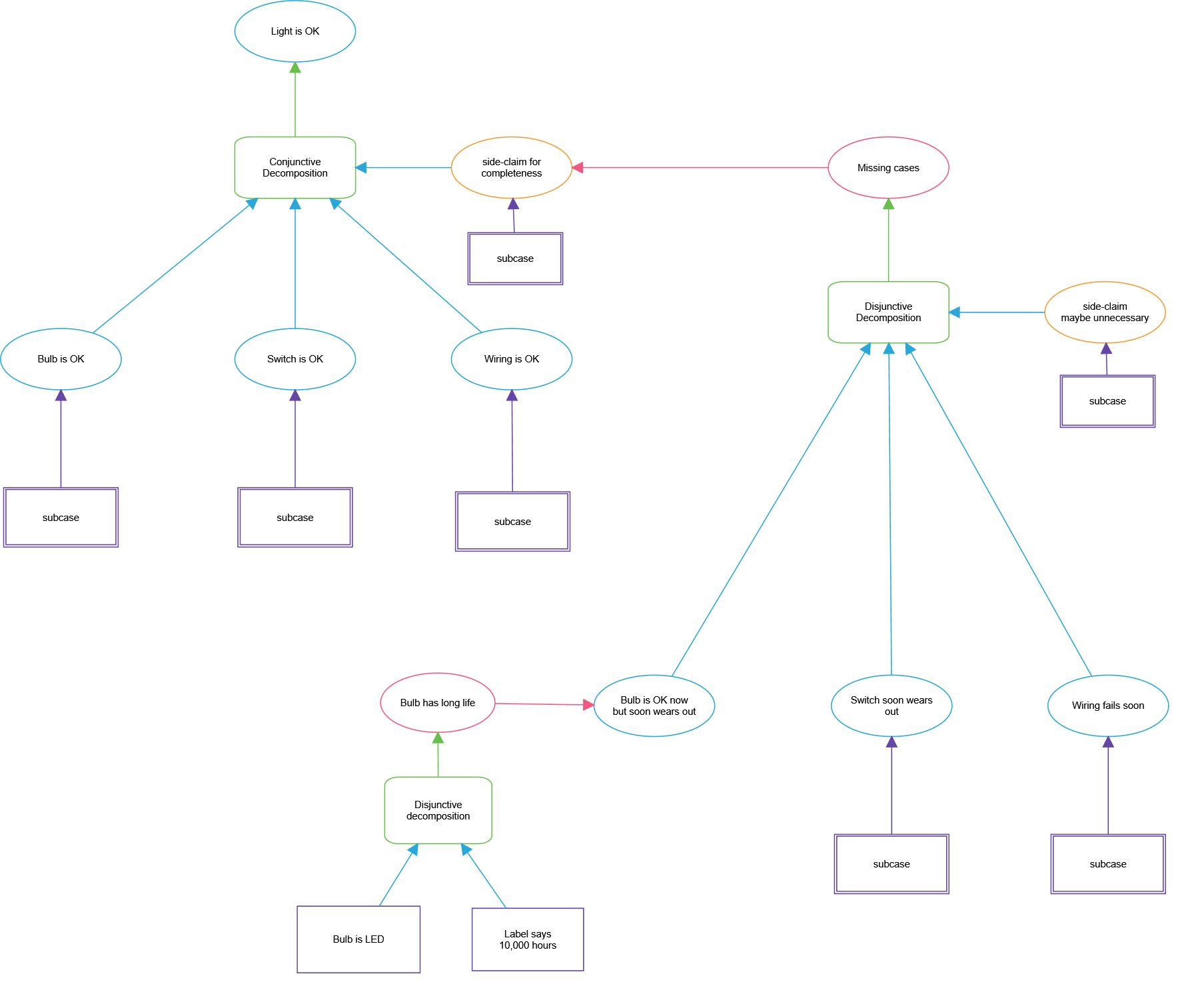}
\end{center}
\vspace*{-3ex}
\caption[Two Levels of Defeaters]{\label{twodef}Two Levels of Defeaters: Exploratory Top Right, Exact Bottom Center}
\end{figure}

The general propagation rules and the use of conjunctive and
disjunctive decomposition blocks, as described in the previous
sections, are illustrated in Figure \ref{twodef}.  Here, we suppose
than an electrician has installed a new light and we want assurance
that it works correctly.  On the left we have a positive case that
decomposes conjunctively into three subcases concerning whether the
light bulb, the switch, and the wiring are OK\@.  For brevity, we omit
concretion and other steps needed to properly connect the top claim to
this decomposition.  We suppose that these subcases all support
\texttt{true} subclaims.  A sideclaim must establish that these are
the only cases that need to be considered and we suppose that its
subcase is also assessed \texttt{true}.  Then, provided the narrative
justification for the decomposition is judged credible, the top claim
can be assessed \texttt{true}.

But on the right, we add an exploratory defeater that introduces
a negative subcase challenging the sideclaim by asserting that the
three cases are insufficient.\footnote{In a more fully developed case,
the defeater would probably challenge a missing concretion step where,
``OK'' is implicitly concreted as ``OK now'' but should surely also
include the expectation that it will continue to be OK for the near
future.}  It would be sufficient to sustain this by an assumption that
states ``working now is not enough: we need to be sure it will
continue to do so, for some specified minimum time.''  Instead, for
the purposes of illustration we suppose that this defeater is
supported by consideration of the possibilities that any of the bulb,
switch, and wiring are OK now, but may soon fail.  This is represented
by a disjunctive decomposition: if any of these can be sustained, then
the defeater is \texttt{true} and the affected sideclaim (and hence
the top claim also) becomes \texttt{unsupported}.

The failure subcases for the switch and wiring are abstracted, but
that for the bulb is developed.  It seems that only ``long life''
bulbs will be used in this installation, so we can refute the claim
that the bulb soon wears out; we do this with an eliminative argument
using an exact defeater that claims the bulb has a long life.   (The
narrative justification for this block should explain why ``Bulb has
long life'' is the negation of ``Bulb is OK now but wears out,''
possibly by refining these claims.)  As a second-level defeater, this
introduces a positive subcase and so we expect any decompositions
below it to be conjunctive.  However, what we wish to assert is that
the bulb used will be new (but ``burned in'') and either an LED, or
one with a label claiming it is good for 10,000 hours.  Both of these
ensure long life and so we use them as subclaims to a disjunctive
decomposition.  As explained in Section \ref{conjdisj}, this is the
only situation we know of where a disjunctive decomposition is
appropriate in a positive case.  The idea is that we wish to provide
assurance for a parameterized system and we disjunctively decompose
the argument into subcases according to different possible
instantiations: the argument serves as a template and in any
particular instantiation only one subcase will be used.  

Here we may suppose that the evidence indicates we have an LED bulb
and we will not use the subcase that requires evidence from the label.
Note that these applications of evidence are abbreviated in the
interests of concision and simplicity; usually, an evidence
incorporation block supports an evidentially measured claim and then a
substitution block is used to lift this to an evidentially useful
claim.

Because the evidence indicates the bulb is an LED, the corresponding
evidential node is \texttt{true} and this propagates through the lower
disjunctive decomposition to make the exact defeater's subclaim
\texttt{true}---and this makes its parent claim \texttt{false}.  This
claim (i.e., ``bulb is OK now but wears out'') is a subclaim to the
upper disjunctive decomposition.  As the other subclaims to this
decomposition have undeveloped subcases, they are assessed
\texttt{unsupported} and the parent claim to the decomposition is
assessed \texttt{unsupported} also (since none of its subclaims are
\texttt{true}).  This parent claim is that in the first-level defeater
and so it propagates as \texttt{unsupported} also and causes the
defeater's target (the sideclaim to the conjunctive decomposition) to
become \texttt{unsupported} and hence the top claim also.  Thus, the
act of investigating this defeater has alerted us that future failure
is a valid concern, and so we will expand the positive case to reflect
this.  We can then mark the defeater as \texttt{addressed}

Note that we developed the subcase to the first, exploratory defeater
in a particular way in order to illustrate exact defeaters,
disjunctive decompositions, and refutational reasoning.  In normal
practice, once we had recognized the need to consider future failures,
we would use an assumption ``light could fail in the near future'' to
sustain the defeater, leading to immediate revision of the case.

The revised case might add a concretion block below the top claim to
define what it means for the light to be OK and this might say that it
works now and can be expected to continue to do so for some specified
time in the future, given a specified pattern of use, and ``expected''
might be further concreted into some probabilistic claim.  The subcase
below that might then decompose into ``now'' and ``future'' branches
and the subcase to the exploratory defeater, suitably adjusted (e.g.,
it becomes a positive case so its disjunctive decomposition is changed
to conjunctive), will become the subcase to the ``future'' branch.

\section{Defeaters and Confidence}

Assurance 2.0 provides two complementary ways for assessing an
assurance case argument and \clasce\ provides automated support for
these.  The first, definitive, assessment is logical indefeasibility;
this requires that the argument is logically valid, the narrative
justifications provide good reasons why each argument block is sound,
and there are no credible reasons that would cause us to revise these
judgments.  We do not use checkmarks or other annotations to record
the judgement that a given block is indefeasibly sound; instead, we
assume this is so and use defeaters to indicate dissent.  Automated
validity checking using the methods developed in the previous sections
takes defeaters, and thereby soundness assessments, into account and
indicates the (in)completeness of the overall argument and the status
of each claim.

It is worth recapitulating our terminology for the various stages of
(in)completeness that an assurance argument may occupy.
\begin{description}
\item[Valid:] this is the standard judgment from logic.  It means that
each claim is supported by an argument block whose subclaims and
sideclaims are likewise supported by further argument blocks,
ultimately terminating in assumptions, evidence, or defeaters
that have been accepted as residual risks.

The graphical interface of \clasce\ is intended to eliminate many
forms of invalidity ``by construction'' and thereby to encourage the
development of arguments that will be valid when completed.

Validity does not depend on the meaning of the claims in an argument:
it is solely concerned with the ``form'' of the argument.  Soundness
takes the meaning into account.

\item[Sound:] again, this judgment comes from logic.  It means that
each argument block has a narrative justification that is considered
(by a human developer or evaluator) to establish that its evidence or
subclaims truly entail the parent claim, given the sideclaim (if any).
For the parent claims of reasoning (i.e., interior) steps the
entailment should be deductive (i.e., have no ``gaps''), and for
evidentially useful claims it is an epistemic judgment that should be
supported by informal or explicit confirmation measures applied to the
evidence.  Likewise, the credibility of assumptions must be supported
by adequate justifications, and defeaters accepted as residual risks
must be supported by narrative justifications that the risk they pose
is tolerable (considering both likelihood and impact).

Again, Assurance 2.0 and its support by \clasce\ are intended to
encourage the development of sound arguments ``by construction'': that
is why it has sideclaims, confirmation measures, and a limited
selection of blocks.

\item[Deductively valid:] soundness sets a high bar; a useful
intermediate step between validity and soundness focuses on whether
the interior argument blocks are \emph{deductive}: that is whether the
subclaims truly entail the parent claim, given the sideclaim (if any).
If an argument block is not deductive, it means (at best) that the
subclaims ``strongly suggest'' the parent claim, and might entail it
given some additional, but presumably unknown, information.  Hence,
there are ``gaps'' between the subclaims that should be filled by
identifying and supplying this absent information (or else the
argument block is completely unsound).  Thus, a deductively valid
argument is one that has no ``gaps'' and a sound argument is a
deductively valid argument that has withstood even stronger scrutiny,
focusing on the details and credibility of its narrative
justifications and confirmation measures.  An alternative to supplying
missing information is to identify it as a residual risk and justify
why it is acceptable.

\item[Indefeasibly sound:] this judgment comes from epistemology.  It
means that our judgement of soundness must be so strong that no
credible new information would change it.

\item[Open:] doubts about an assurance argument are indicated by
attaching (exploratory) defeaters.  Defeaters that are themselves
defeated (i.e., are refuted by a subargument) are said to be
``retired,'' meaning they no longer play a part in the assurance
argument, but may be retained as commentary.  Unrefuted or true
defeaters may be explicitly accepted and justified as residual risks;
otherwise they are said to be ``active.''  An assurance argument that
has active defeaters or is incomplete (e.g., has disconnected
subarguments or has unsupported claims) is said to be ``open,''
otherwise it is ``closed.''
\end{description}

The validity plugin of \clasce\ checks the validity of closed
arguments and provides helpful feedback on open ones.

The use of defeaters and refutational reasoning within validity
assessment supports a cooperative balance between human judgment and
automation in assessment of soundness and indefeasibly: formulating
defeaters, and exploring their subcases, expresses human judgment
about local soundness, while validity assessment incorporating
refutational reasoning integrates these judgments over the full case.

The second form of assessment for assurance case arguments concerns
confidence and, in its strict form, this applies only to arguments
that have already been assessed as indefeasible, or at least sound.
Now, it is reasonable to ask how we can have anything less than full
confidence in an argument that is indefeasibly sound.  The explanation
is that our judgements of soundness---that is, whether each evidential
block supports its evidentially useful claim and whether the subclaims
of each reasoning block indefeasibly entail their parent claim (given
any sideclaims)---are human judgments, and confidence, quantified as a
subjective probability, indicates our \emph{degree of belief} in that
judgement, which may be less than total even for indefeasible claims
(e.g., the judgment of indefeasibility may rest on hazard analysis,
which can never be declared perfect).

We want to assess confidence compositionally: that is, confidence in a
(sub)case should be calculated from confidence in its parts and this
requires that confidence is represented numerically, so that we can do
arithmetic with it.\footnote{We could use some ordinal scale, such as
low medium, and high, or even acceptable and unacceptable, and then
devise rules for their combination, but there seems little merit in
doing so.}  The
natural way to do this is to represent confidence as a subjective
probability.

For evidence, confidence is derived from the quantitative or
qualitative assessments used in calculating confirmation measures for
the evidentially useful claim: usually, $P(C \vbar E)$ is used as the
confidence measure, that is, the posterior likelihood of the
evidentially useful claim, given the evidence.  For other blocks, the
default assessment is derived by a ``sum of doubts'' calculation,
where doubt is the dual of confidence (i.e., $\emph{doubt} = 1 -
\emph{confidence}$), so the doubt in a parent claim is given by the
sum of the doubts in its subclaims and sideclaim.  This default
assessment can be modified by the user to reflect human judgment.  The
\clasce\ confidence plugin propagates these human and automated local
assessments throughout the case.

Confidence assessments of completed cases are not used to deliver
judgment on a case (that is the r\^{o}le of indefeasible soundness)
but to help ensure balanced effort across a case and to support
graduated assessments.  Graduated assessments assist in trading
confidence for cost in applications considered to pose less risk, as
exemplified by the Design Assurance Levels (DALs) of DO-178C
\cite{DO178C} and System Integrity Levels (SILs) of IEC 61508
\cite{IEC61508}.  Cost can be reduced by simplifying the system and
thereby its assurance case (e.g., providing less fault checking or
redundancy), and/or by weakening the assurance case (e.g., providing
less evidence, or less costly evidence).  It is also possible that
system-level assurance cases may employ different levels of confidence
in different parts of the case.  For example, an architectural
framework that enforces partitioning \cite{Rushby99:partitioning} may
require the highest level of confidence in its assurance, while lesser
levels of confidence may be adequate for some of its partitioned
applications (because partitioning limits fault propagation).

Logical validity and confidence are evaluated in \clasce\ by separate
plugins and both of these can decorate the graphical representation of
the case to indicate their assessments.\footnotet{The \clasce\
validity plugin does not currently decorate the graphical
representation.}  Logical validity uses the propagation rules of
Section \ref{ref} (as augmented by Section \ref{conjdisj} if
disjunctive decompositions are present).  Confidence is assessed by
the methods sketched above and described in detail in our Confidence
Report \cite[Section 3]{Bloomfield&Rushby:confidence21}.

Strictly, both logical and confidence assessments apply only to closed
cases: that is, those in which there are no unsupported claims nor
unresolved doubts nor active exploratory defeaters.  Nonetheless, it
can be helpful to calculate approximations to these assessments for
open cases during construction or assessment (e.g., when assessors
have used defeaters to flag or explore doubtful elements of a case),
either to estimate progress or to focus attention, and the \clasce\
plugins can do this.

We suggest using \clasce\ and its plugins to support these activities
as follows.  Logical validity should be checked frequently during
development; in the early stages, it is expected that most parts of
the argument will be assessed \texttt{unsupported}, with only some
areas assessed \texttt{true}.  Temporary assumptions can be used for
the purpose of exploration: for example, in the absence of some
evidence we could assert the intended evidentially useful claim as an
assumption in order to check that a valid case can be built above it.
Negative assumptions (i.e., assessments of \texttt{false}) can be
achieved by adding an exact defeater above a positive
assumption.
Narrative justifications can be supplied either as the argument is
developed, or once a valid skeleton for it is in place.
Deductiveness, soundness, and indefeasibility can then be considered,
either serially or all together, and any doubts about these can be
recorded and explored by adding defeaters to the argument and
exploring their impact using the validity checker.

Confidence assessments are typically propagated upward from evidence
and assumptions using the sum of doubts calculation, and are performed
only when the argument is moderately complete (i.e., has no
disconnected subarguments nor unsupported claims).  Confidence values
can be adjusted by the human user: the purpose of in-progress
confidence assessments is not to deliver judgement but to assist
allocation of effort and to direct attention to those parts of an
assurance argument most in need of attention.

\section{Summary and Conclusion}

We have introduced two ways of using defeaters in \atwo
\cite{Bloomfield&Rushby:Assurance2} and  its \clasce\ toolset
\cite{Varadarajan-all:DASC24}.  Ordinary, or \emph{exploratory}
defeaters provide a way of recording and exploring doubts about an
assurance case argument or its narrative justifications and can be
retained as commentary to assist future developers and evaluators who
may have similar doubts.  Exact or \emph{eliminative} defeaters
introduce negation into an argument: that is, instead of providing a
subcase sustaining claim $A$, we instead attempt to provide one that
refutes $\neg A$.  Both kinds of defeater help challenge confirmation
bias by inviting consideration of a contrary point of view in the
manner of a dialectic or Socratic dialog.  Exact defeaters allow an
assurance case or subcase to proceed by \emph{eliminative
argumentation}, which some find more persuasive than a conventional,
positive assurance case.

By a \emph{positive} assurance case or subcase we mean one that
employs an argument where the assurance goal is to \emph{sustain} the
local top (sub)claim (i.e., to assess it as \texttt{true}); the
alternative is a \emph{negative} case, where the goal is to
\emph{refute} the local top subclaim (i.e., to assess it as
\texttt{false}).  The subcase to any first-level defeaters of a
positive case will be negative and vice-versa.  Defeaters may appear
in the subcases of other defeaters so these assessments can alternate.
Thus, a subargument is positive if it is under an even (or zero)
number of defeaters and negative if it is under an odd number.

The claims of an assurance argument may be assessed as \texttt{true},
\texttt{false}, or \texttt{unsupported} and these assessments
propagate in suitable ways upward from the leaf nodes of the argument
(i.e., from evidence, assumptions, and residual risks).  Defeaters
must be treated suitably and, in particular, care is needed when
propagating \texttt{false} as this can introduce the fallacy of
\emph{denying the antecedent}.  These concerns, and their correct
treatment in \clasce, are described in detail in Section \ref{alg}.

Truth assignments to claims propagate the same in positive and
negative cases and there is no difference in interpretation or meaning
between positive and negative cases, it is just that we usually try to
sustain the former and refute the latter.  In a finished assurance
case argument, all positive cases should be sustained and all negative
ones refuted (or, exceptionally, accepted as \emph{residual risks}
\cite[Section 6]{Bloomfield&Rushby:confidence21}) as otherwise some
defeaters must still be unresolved (i.e., the case is still \emph{open}).

There is one difference between positive and negative arguments:
subclaims to decomposition blocks are typically conjoined in positive
arguments, but disjoined in negative ones.  We introduce disjunctive
decomposition blocks to \clasce\ so that this choice can be
represented explicitly.  This also allows disjunctive decompositions
to be used in positive arguments (and conjunctive in negative ones)
and Section \ref{conjdisj} examined the (rare) circumstances where
this can be useful.

When the claim of an exploratory defeater is refuted, it means that
the doubt that motivated it is annulled and the original case stands;
however, the defeater and its subargument can be retained as
commentary.  If refutation is unsuccessful, then we can attempt
instead to sustain the defeater's claim (this is an instance where we
attempt to sustain a negative subcase and will generally require
adjustment to the subcase).  If this succeeds, it means that the doubt
is justified and the primary argument, and possibly the system itself,
must be revised.  Once that has been accomplished, it should be
possible to refute the defeater and to retain it and its (likely once
again adjusted) subcase as commentary.

The claim in an \emph{exact} defeater must be the negation of the
claim or defeater that it points to, so that refutation of the
defeater claim sustains the target claim and vice-versa.  Because
claims expressed in natural language can be hard to analyze (to see if
one is the negation of another), users can direct \clasce\ to mark
defeaters as exact, in which case the subclaim and parent claim are
assumed to be negations of each other, and any existing subcase for
the affected claim is temporarily ignored.  Those who favor eliminative argumentation can
employ it by simply introducing an exact defeater near the top of the
argument, while those who do not approve this style of argumentation
can eschew this construction.

Refutational arguments, as introduced by defeaters, invite a different
perspective than conventional positive arguments and can be valuable
in challenging confirmation bias and other forms of complacency in the
construction of assurance cases.  However, the finished assurance case
delivered to evaluators must be indefeasibly sound (modulo residual
risks that are documented and explicitly accepted), so all exploratory
defeaters must have been refuted and no longer play an active part in
the argument.  Nonetheless, consideration of refuted defeaters can
greatly assist evaluators (and future (re)developers) to comprehend a
case, and \ifelse{\clasce}{the \clasce\ tool for \atwo} can therefore
retain them and their subcases in completed cases, where they function
as a kind of commentary.  To avoid cluttering the main argument, and
also to accommodate those who consider that evaluators should review a
finished argument independently of its developers,
\clasce\ can selectively hide  or reveal
defeaters and their subcases.

Evaluation of an assurance case and, on that basis, authorizing
deployment of the system concerned, are topics of delicate judgement.
It is not the task of the evaluators to repeat the work of the
assurance case developers, but they must form some judgement about it
and there is contention whether defeaters should play a part in this.
Our colleague Shankar says: ``an assurance argument is a brief, not a
debate'' but, on the other hand, evaluators may want to know that
developers have considered contrary points of view, and they may have
doubts of their own and will want to see if these were considered and
how they were resolved.  We refer to this process of review and
assessment as the ``case about the case'' or \emph{metacase} and are
preparing a report on the topic.  We are also preparing a report on
systematic methods of searching for potential defeaters.

In summary, defeaters and the refutational (sub)arguments that they
introduce are vital tools in the construction of sound and persuasive
assurance case arguments, and potentially also in their assessment.
However, their use and assessment (particularly at multiple levels) is
not straightforward, and we hope that this report has adequately
explained and justified their treatment in \ifelse{\clasce}{\atwo\ and
its \clasce\ tool}.

\paragraph{Acknowledgments.}

The work described here was developed in partnership with other
members of the \cl\ project, notably Srivatsan Varadarajan, Anitha
Murugesan, and Isaac~Hong Wong of Honeywell, Gopal Gupta of UT Dallas,
and Robert Stroud of Adelard.

This material is based upon work performed under subcontract to
Honeywell supported by the Air Force Research Laboratory (AFRL) and
DARPA under Contract No.\ FA8750-20-C-0512\@.  Any opinions, findings
and conclusions or recommendations expressed in this material are
those of the author(s) and do not necessarily reflect the views of the
Air Force Research Laboratory (AFRL) and DARPA.

\addcontentsline{toc}{section}{References}
\bibliographystyle{modplain}

\end{document}
% End of foo.tex -----------------------------------